\documentclass{article}

\PassOptionsToPackage{numbers}{natbib}
\usepackage[preprint]{neurips_2021}

\usepackage[utf8]{inputenc} 
\usepackage[T1]{fontenc}    
\usepackage{hyperref}       
\usepackage{url}            
\usepackage{booktabs}       
\usepackage{amsfonts}       
\usepackage{nicefrac}       
\usepackage{microtype}      
\usepackage{xcolor}         
\usepackage{comment} 
\usepackage{amsmath} 
\usepackage{amssymb} 
\usepackage{graphicx}
\usepackage{caption}
\usepackage{subcaption}
\usepackage{tikz}
\usepackage[normalem]{ulem}

\newcommand{\x}{\boldsymbol{x}}
\newcommand{\y}{\boldsymbol{y}}
\newcommand{\z}{\boldsymbol{z}}
\newcommand{\el}{\boldsymbol{l}}

\newcommand{\de}{\boldsymbol{d}}

\newcommand{\srFL}[1]{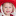} 
\newcommand{\srFB}[1]{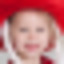} 
\newcommand{\srFE}[1]{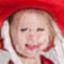} 
\newcommand{\srFS}[1]{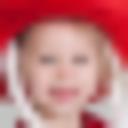} 
\newcommand{\srFP}[1]{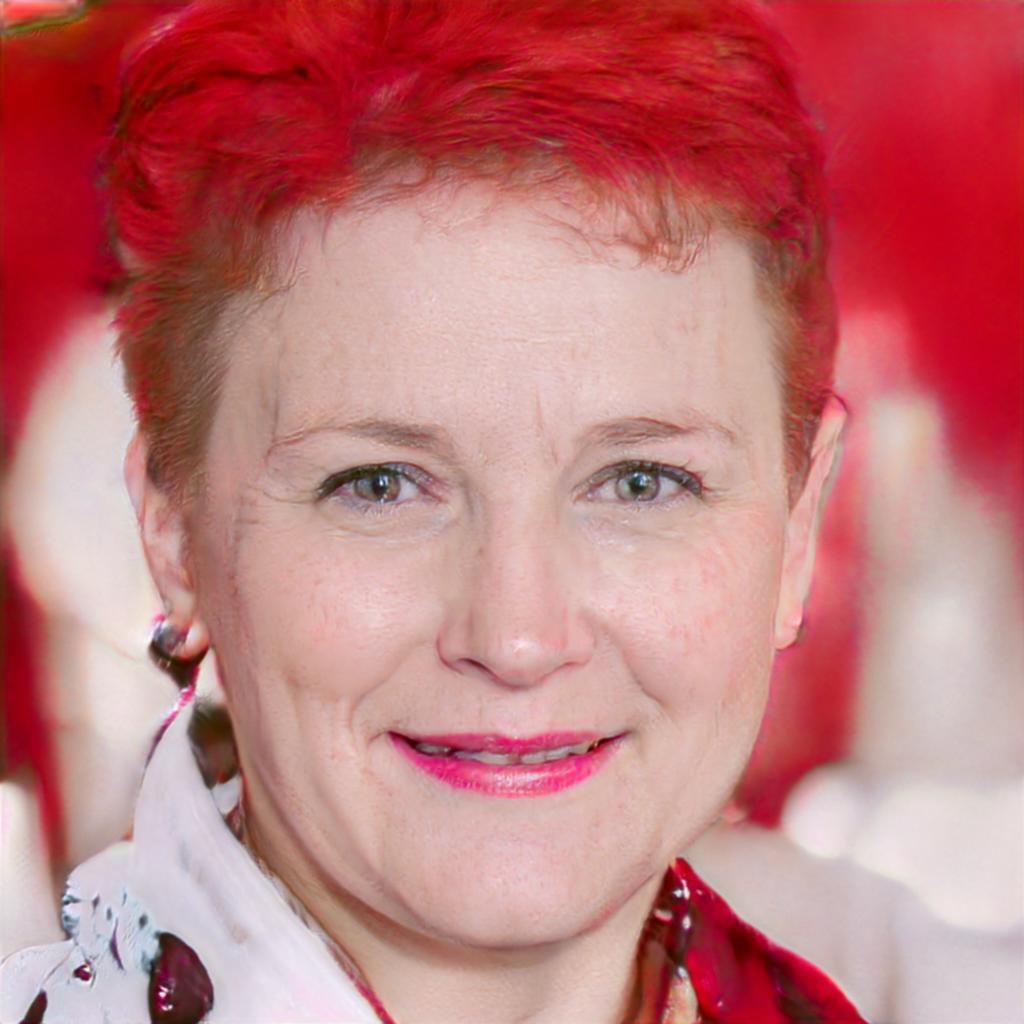} 
\newcommand{\srFO}[1]{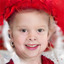} 
\newcommand{\srFJ}[1]{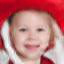} 
\newcommand{\srFH}[1]{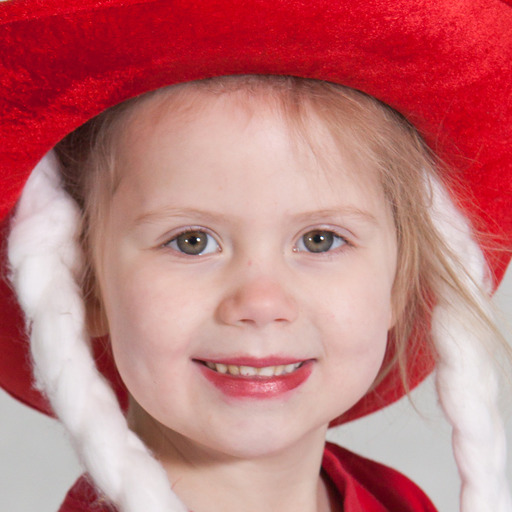} 

\newcommand{\inc}[1]{\includegraphics[height=\w]{#1}}
\newcommand{\w}{1.7cm}

\title{CAFLOW: Conditional Autoregressive Flows}

\author{%
  Georgios Batzolis\\
  DAMTP\\
  University of Cambridge\\
  Cambridge CB3 0WA\\
  \texttt{gb511@cam.ac.uk} \\
  \And
  Marcello Carioni\\
  DAMTP\\
  University of Cambridge\\
  Cambridge CB3 0WA\\
  \texttt{mc2250@cam.ac.uk} \\
  \And
  Christian Etmann\\
  DAMTP\\
  University of Cambridge\\
  Cambridge CB3 0WA\\
  \texttt{ce377@hermes.cam.ac.uk} \\
  \AND
  Soroosh Afyouni\\
  Department of Psychology\\
  University of Cambridge\\
  Cambridge CB2 3EB\\
  \texttt{srafyouni@gmail.com} \\
  \And
  Zoe Kourtzi \\
  Department of Psychology\\
  University of Cambridge\\
  Cambridge CB2 3EB\\
  \texttt{zk240@cam.ac.uk} \\
  \And
  Carola-Bibiane Schönlieb\\
  DAMTP\\
  University of Cambridge\\
  Cambridge CB3 0WA\\
  \texttt{cbs31@cam.ac.uk}
}

\begin{document}

\maketitle

\begin{abstract}
  
    We introduce CAFLOW, a new diverse image-to-image translation model that simultaneously leverages the power of auto-regressive modeling and the modeling efficiency of conditional normalizing flows. We transform the conditioning image into a sequence of latent encodings using a multi-scale normalizing flow and repeat the process for the conditioned image. We model the conditional distribution of the latent encodings by modeling the auto-regressive distributions with an efficient multi-scale normalizing flow, where each conditioning factor affects image synthesis at its respective resolution scale. Our proposed framework performs well on a range of image-to-image translation tasks. It outperforms former designs of conditional flows because of its expressive auto-regressive structure.

\end{abstract}

\section{Introduction}

Generative modeling has emerged as one of the most widely-researched areas in deep learning over the last few years. While generative adversarial networks (GANs) \cite{GANs} produce state-of-the-art results for images \cite{viazovetskyi2020stylegan2}, they do not allow for the estimation of likelihoods. Other types of generative methods, however, admit the explicit evaluation of likelihoods of points under the given model, and thus allow for training as maximum likelihood estimators. Normalizing flows \cite{rezende2015variational} (and their continuous formulation, via NeuralODEs \cite{neuralODEs}) are one such example. These \emph{flow-based} models are diffeomorphic neural networks, which are trained to invertibly transform data into e.g. a normal distribution. This is made possible by a change-of-variables formula, which expresses the likelihood of a data sample in terms of the normal distribution. Samples from the approximated distribution are then generated by passing samples from a normal distribution through the inverse function.
A straightforward extension of likelihood estimators are \emph{conditional} likelihood estimators, which represent the likelihood of a random variable conditioned on another random variable. This is in particular used in autoregressive models and their flow-based variants (\emph{autoregressive flows} \cite{autoregressive_flows}). These types of models explicitly parametrize joint likelihoods via the product rule for probability densities, and can hence be used for creating expressive likelihood estimators, at the cost of more involved computations. This approach is in particular employed by Wavelet Flows \cite{WAVELET-FLOW}, which use a hierarchical multi-scale representation via Wavelet decompositions in order to parametrize expressive conditional likelihood functions for images.

Domain transfer, e.g. via models such as CycleGAN \cite{CycleGAN2017} and DUAL-GLOW \cite{dual-glow}, is another topic that has received wide interest in recent years. In domain transfer, the task is to \emph{transfer} a data point from some \emph{origin} domain (e.g. black-and-white images) to a different \emph{target} domain (color images in this example). The technique of conditioning lends itself well to domain transfer, as it is possible to condition likelihood functions over target domain on points from the origin domain, thereby making conditional generation possible. In flow-based models, this has been introduced in the form of conditional flows \cite{cflows}, which function based on this general principle. SRFlow \cite{SRFLOW} and DUAL-GLOW can be seen as special cases of this, where the conditioning on the origin domain is done in a multi-scale fashion.

\subsection{Contributions}
In this work, we introduce CAFLOW, \emph{conditional autoregressive flows}, which combine the idea of conditional flows with the expressiveness of autoregressive models using hierarchical multi-scale flows for domain transfer. 
As a first step, our model encodes the conditioning and the conditioned images into a hierarchical sequence of latent spaces. Then, using an autoregressive approach, it decomposes the conditional distribution of the latent encodings into $n$ autoregressive components with the aim of modeling each component using a conditional normalizing flow.
By carefully designing such a flow based on weak assumptions on the mutual dependencies of the latent encodings, our model allows for exchange of information between latent spaces of different dimension. 
In particular, our designed architecture is able to capture correlations between different scales, improving on the expressivity of other conditional flow-based models used for domain transfer tasks.
To the best of our knowledge, our work is the first to explore cross-scales correlations in the context of conditional normalizing flows for domain adaptation.  

Our modeling choices are corroborated by strong experimental evidence. We demonstrate that our model achieves good results on classical image-to-image translation tasks such as image super-resolution, image colorization and image inpainting. In particular, we show that on these tasks CAFLOW outperforms former designs of conditional flows and GAN-based models thanks to its expressive auto-regressive structure.

\section{Background}\label{sec:background}

\subsection{Normalizing flows}\label{subsec:normalizing}

Given a random variable $Y$ with an unknown distribution $p_Y$, the main idea of flow-based generative modeling is to approximate $p_Y$ with a learned distribution $p^\theta_Y$, which is parametrized by an invertible neural network $G^\theta$ mapping from a latent space $Z$ to $Y$. 
Choosing $p_Z$ to be a tractable distribution on $Z$, we define the distribution $p^\theta_Y$ as the one obtained by applying the mapping $\y = G^\theta(\z)$ to the distribution $p_Z$.
By the change of variables formula for probability distributions, $p^\theta_Y$ is easily computed as
\begin{align*}
p^\theta_Y(\y) = p_Z(F^\theta(\y)) \left| \det \left(\frac{d F^\theta(\y)}{d\y} \right)\right|,
\end{align*}
where $F^\theta$ is the inverse of $G^\theta$.
Thanks to the formula above it is then possible to match $p^\theta_Y$ to $p_Y$ by miminizing the negative log-likelihood of $p^\theta_Y$,
\begin{align*}
- \log p^\theta_Y(\y) = -  \log(p_Z(F^\theta(\y))) -  \log\left| \det \left(\frac{d F^\theta(\y)}{d\y} \right)\right|.
\end{align*}
In practice, the transformation $G^\theta$ is a composition of learnable invertible transformations $g_{1}, ..., g_{n}$ such that $g_i : Z_{i} \rightarrow Z_{i+1}$ (where $Z_1 = Z$ and $Z_{n+1} = Y$) for intermediate latent spaces $Z_i$ and
\begin{equation}
    \y = g_{n}(g_{n-1}(...g_{1}(\z)))=G^{\theta}(\z)\,, \quad  \z= f_1(f_{2}(...f_n(\y)))=F^{\theta}(\y),
\end{equation}
where $f_i = (g_i)^{-1}$ for every $i$ and thus $F^\theta = (G^\theta)^{-1}$. This model is called a normalizing flow.
In this case, using the properties of the Jacobian for composition of invertible transformations the log-likelihood of $p^\theta_Y$ can be computed as 
\begin{equation}
\log(p^\theta_Y(\y) ) = \log(p_Z(F^\theta(\y))) + \sum_{i=1}^{n}  \log\left| \det \left(\frac{d{f_i}(\z_{i+1})}{d\z_{i+1}} \right)\right|,
\end{equation}
where $\z_{n+1} = \y$.
In order to estimate $\log(p^\theta_Y(\y) )$ the transformations $g_i$ need to be designed to have a computable inverse $f_i$ and a tractable Jacobian determinant.
We next describe typical choices for the transformations $g_i$ introduced in \citep{GLOW, Nice2014} and then commonly used in most of the normalizing flow architectures. We will refer to them in the next sections where we describe our method.

\textbf{Affine coupling layer.} Affine coupling layers \citep{Nice2014} capture complex dependencies of the activations in an invertible way. They split the activation in the dimension of the channels creating two components $\z_1$ and $\z_2$. $\z_2$ is untouched by the transformation and is used to calculate the scale and the bias in the affine transformation of $\z_1$ as shown in \eqref{eq:affinecoupling}. The scale and the bias are calculated by parametrized neural networks $h_s$ and $h_b$ which do not need to be invertible. 
\begin{align}\label{eq:affinecoupling}
& \y_1 = \exp(h_s(\z_2))\z_1 + h_b(\z_2)\\
& \y_2 = \z_2 \nonumber
\end{align}
The output is then concatenated in $[\y_1,\y_2]$. Note that the inverse and the Jacobian determinant of the transformation \eqref{eq:affinecoupling} can be efficiently computed \citep{Nice2014}.

\textbf{Invertible $1 \times 1$ convolution.} Invertible $1 \times 1$ convolutions have been introduced in GLOW \citep{GLOW}. They replace standard convolutional layer for which inverse and Jacobian determinant are not tractable. They mix the activations in the dimension of the channels so that different parts of the activation are modified by subsequent coupling layers.

\textbf{Actnorm.} Actnorm has been introduced in \citep{GLOW} as a variant of the classical batch normalization \citep{batchnormalization} where the minibatch size is equal to one. 
It performs an affine transformation using a scale and bias parameter per channel. It improves the training stability.

\subsection{Conditional Normalizing Flows}

The normalizing flow approach can be adapted to model conditional densities of complicated target distributions. Precisely, a conditional density $p_{Y|X}$ is parametrized using a transformation $G^\theta : Z \times X \rightarrow Y$ such that $G^\theta(\cdot, \x) : Z \rightarrow Y$ is invertible for every condition $\x$. We denote by $p^\theta_{Y|X}$ the distribution obtained by applying the mapping $\y = G^\theta(\z, \x)$ to samples $\z$ from a simple distribution $p_Z$. By the change of variables formula for probability distributions the conditional log-likelihood of $p^\theta_{Y|X}$ can be then computed as 
\begin{align}\label{eq:cond}
\log(p^\theta_{Y|X}(\y|\x)) = \log(p_{Z}(F^\theta(\y,\x)))   +  \log\left| \det \left(\frac{\partial F^{\theta}(\y,\x)}{dy} \right)\right|,
\end{align}
where $F^\theta(\y, \x)$ is the inverse of $G^\theta(\z, \x)$ for every condition $\x$.
Consequently, a generative model for $p_{Y|X}$ can be trained by minimizing the negative log-likelihood of the parameters $\theta$ using the formula in \eqref{eq:cond}. The sampling procedure works similarly to standard normalizing flows. We generate a sample $y \sim p_{Y|X}$ by sampling a latent $\z\sim p_Z$ and passing it through $G^\theta(\cdot, \x)$, yielding $y=G^\theta(z, \x)$. Similarly to traditional normalizing flows the map $G^\theta$ is modelled through a composition of learnable invertible conditional transformations. We describe next the specific layers used in conditional normalizing flow architectures.

\textbf{Conditional affine coupling layer.} It is a conditional variant of the affine coupling layer. The difference is that the scale and the bias are calculated by parametrized functions of both the slit activation $\z_2$ and the condition $\x$ as shown in \eqref{eq:conditional affine coupling layer}.
\begin{equation}
 \begin{aligned}\label{eq:conditional affine coupling layer}
\y_1 &= \exp(h_s(\z_2,\x))\z_1 + h_b(\z_2,\x)\\
\y_2 &= \z_2 
\end{aligned}   
\end{equation}

\textbf{Affine Injector.} The affine injector layer has been introduced in \citep{SRFLOW} to transfer more information from the conditioning data to the main branch of the flow. It is an affine transformation of the flow activation $\z$, where the scale and the bias are parametrized functions of the condition $\x$.
\begin{align*}
\y =  \exp(h_s(\x))\z + h_b(\x).
\end{align*}
The inverse and the Jacobian determinant of such transformation can be easily computed as in \citep{SRFLOW}.

\section{Related work}

Domain adaptation and image-to-image translation tasks have received a lot of attention in recent years, thanks to the development of representation learning methods \citep{representationsurvey} and the advent of adversarial approaches \citep{GANs, WGAN}. A plethora of methods based on GANs have been proposed for solving super-resolution, colorization, inpainting and many other domain adaptation tasks \citep{CycleGAN2017, colorGAN, cycada, Pixel-leveldomain, ESRGAN, BRGM, Generativecontexualattention, coupled, abdal2019image2stylegan} achieving remarkable results. Along with GAN-based models, other alternatives have been developed: Variational Autoencoders \citep{VAE} approaches \citep{yan2016attribute2image}, score-based models \citep{scorebased} and more in general models that use specific discrepancies to measure the similarity between source and target distributions \citep{Rozantsevbeyond, Sunfrustratingly, BRGM, li2019feedback,  pan2010domain}. 
Conditional normalizing flows and their multi-scale variants have received much less attention for solving domain adaptation tasks. However, in recent years, they have gained popularity. Designed to match source and target distributions by maximizing the likelihood of a parametrized family of probabilities \citep{GLOW, RealNVP2016, cflows, rezende2015variational} they have been used to achieve comparable performances to GAN-based models in super-resolution \citep{SRFLOW, cflows}, inpainting \citep{cGLOW} and image-to-image translation \citep{Pumarola2020, dual-glow, grover2020alignflow}.
Autoregressive models and autoregressive flows are considered one of the current state-of-the-art architecture for image modeling and they have been employed successfully for various tasks, such as image generation, image completion and density estimation \citep{autoregressive_flows, HPNAF, PRRN, larochelle2011neural, germain2015made, NIPS2016_ddeebdee} 

\subsection{DUAL-GLOW}
Here we describe in more detail the DUAL-GLOW model introduced in \citep{dual-glow}, as it shares some similarities with our approach. DUAL-GLOW proposes to couple two multi-scale normalizing flows modelled with a GLOW architecture that interact at different scales. Their main goal is to translate MRI images to PET images, but the same architecture has proven to be useful for other standard image-to-image translation tasks.
They estimate the conditional probability $P(W|Y)$ using two multi-scale normalizing flows which convert images $Y$ and $W$ into their respective hierarchical latent spaces $[D_{n-1}, ..., D_0]$ and $ [L_{n-1}, ..., L_0]$, as shown in schematically in Figure \ref{fig:high_level_design_conditional}. They model the conditional distribution $P(L_{i}|D_{i})$ by a parametrized multidimensional Gaussian distribution whose mean $\mu^{\theta}$ and covariance $\Sigma^{\theta}$ are learnable functions of $D_{i}$. Then they use $P(L_{i}|D_{i})$ to estimate $P(L_{n-1}, ..., L_0|D_{n-1}, ..., D_0)$ under the implicit assumption that information is exchanged only between  latent spaces of the same dimension, that is
\begin{align}\label{eq:dual-glowdep}
P(L_{n-1}, ..., L_0|D_{n-1}, ..., D_0)  = \prod_{i=0}^{n-1}P(L_{i}|D_{i}).
\end{align}

\section{Method}\label{sec:Caflow}
The aim of our CAFLOW model is to perform image-to-image translation tasks by learning the conditional distribution $P(W|Y)$ where $W$ and $Y$ are random variables that model given image distributions.
Following a similar architecture to \citep{dual-glow} we use two multi-scale normalizing flows $R^\theta$ and $T^\theta$ to convert images $Y$ and $W$ into two sequences of $n$ hierarchical latent spaces of decreasing dimension: \(R^\theta(Y)=\tilde{D}_n\), where \(\tilde{D}_n:=[D_{n-1}, ..., D_0]\) and \(T^\theta(W)=\tilde{L}_n\), where \(\tilde{L}_n := [L_{n-1}, ..., L_0]\). Then we design an autoregressive model based on conditional normalizing flows to learn the conditional distribution $P(\tilde{L}_n| \tilde{D}_n)$.
\subsection{Modeling assumptions}\label{subsec:modass}

\begin{figure}[t]
\centering
   \begin{subfigure}{0.3\textwidth}
		\includegraphics[width=\textwidth]{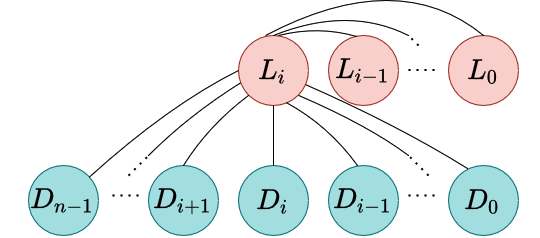}
	\end{subfigure}%
	\quad
\begin{subfigure}{0.3\textwidth}
	\includegraphics[width=\textwidth]{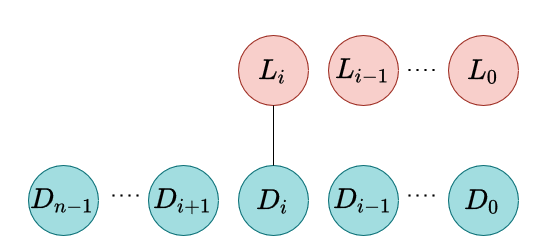} 
\end{subfigure}%
\quad 
\begin{subfigure}{0.3\textwidth}
	\includegraphics[width=\textwidth]{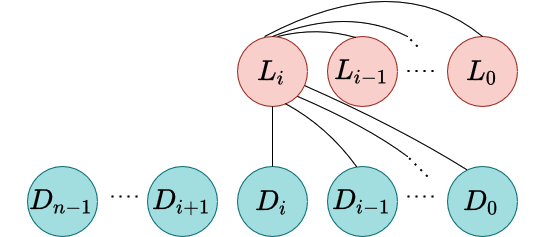}
	\end{subfigure}
\caption{From left to right: ideal dependencies in the $i^{th}$ autoregressive component. DUAL-GLOW modeling assumption \citep{dual-glow}; information is exchanged only between latent spaces having the same dimension. Our modeling assumption; we retain the dependencies between $L_i$ and the latent spaces of lower dimension.}	\label{fig:dependencies}
\end{figure}

The conditional distribution $P(\tilde{L}_n| \tilde{D}_n)$ is factorized using the chain rule of probability into $n$ autoregressive component distributions as shown in \eqref{conditional distribution autoregressive factorization}:
\begin{equation}\label{conditional distribution autoregressive factorization}
    P(\tilde{L}_n| \tilde{D}_n) = \prod_{i=0}^{n-1}P(L_i|\tilde{L}_{i},D_{n-1},...,D_0)
\end{equation}
with the notational convention that $\tilde{L}_{0}=\emptyset$ and $\tilde L_{i} = [L_{i-1}, \ldots, L_0]$ for every $i$. The dependencies of the $i^{th}$ component distribution are shown graphically in the left diagram of Figure \ref{fig:dependencies}. Modeling those $n$ autoregressive distributions can be unnecessarily computationally expensive. For this reason, we assume that 
\begin{equation}\label{eq:factorization}
    P(\tilde{L}_n| \tilde{D}_n)= \prod_{i=0}^{n-1}P(L_i|\tilde{L}_{i},\tilde D_{i+1}),
\end{equation}
where $\tilde D_{i+1} = [D_{i}, \ldots, D_0]$. In particular, we retain the dependencies between $L_{i}$ and all $L$ and $D$ latent variables of level $i$ and below, which effectively means that we are pruning only the dependencies between  $L_{i}$ and $D_{i+1},...,D_{n-1}$. We advocate that this is a valid assumption, because the encoded information in the split variables of the multi-scale flow typically ranges from local noise patterns and image details to higher-level information as we move from the early split variables to the final split variables. 
We remark that our modeling assumption is weaker than the one implicitly used in \citep{dual-glow} described in \eqref{eq:dual-glowdep} (see Figure \ref{fig:dependencies}). Precisely we allow information to be exchanged between latent spaces of different dimension. As we will demonstrate in the experiments our choice allows for more expressive architectures able to capture correlations between different scales.
We represent schematically in Figure \ref{fig:dependencies} the difference between the theoretical latent space dependencies, the one assumed in \citep{dual-glow} and our modeling choice. 
Under this modeling assumption, our goal is to estimate the conditional distributions $P(L_i|\tilde{L}_{i},\tilde D_{i+1})$ for every $i$ and then recover 
$P(\tilde{L}_n| \tilde{D}_n)$ using \eqref{eq:factorization}.

\subsection{Modeling the autoregressive components using conditional normalizing flows}\label{Autoregressive conditional flow components}

\begin{figure}[t]
\centering
\begin{subfigure}{0.45\textwidth}
	\includegraphics[width=\textwidth]{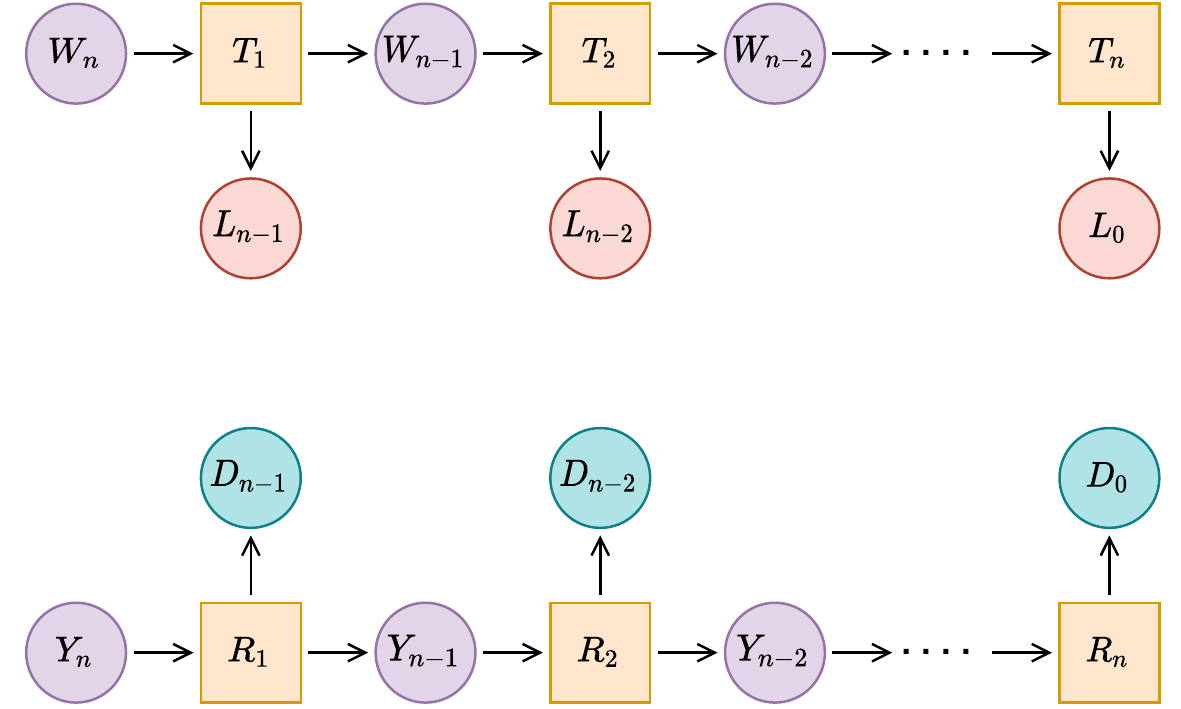}
\end{subfigure}%
\qquad 
\begin{subfigure}{0.47\textwidth}
		\includegraphics[width=\textwidth]{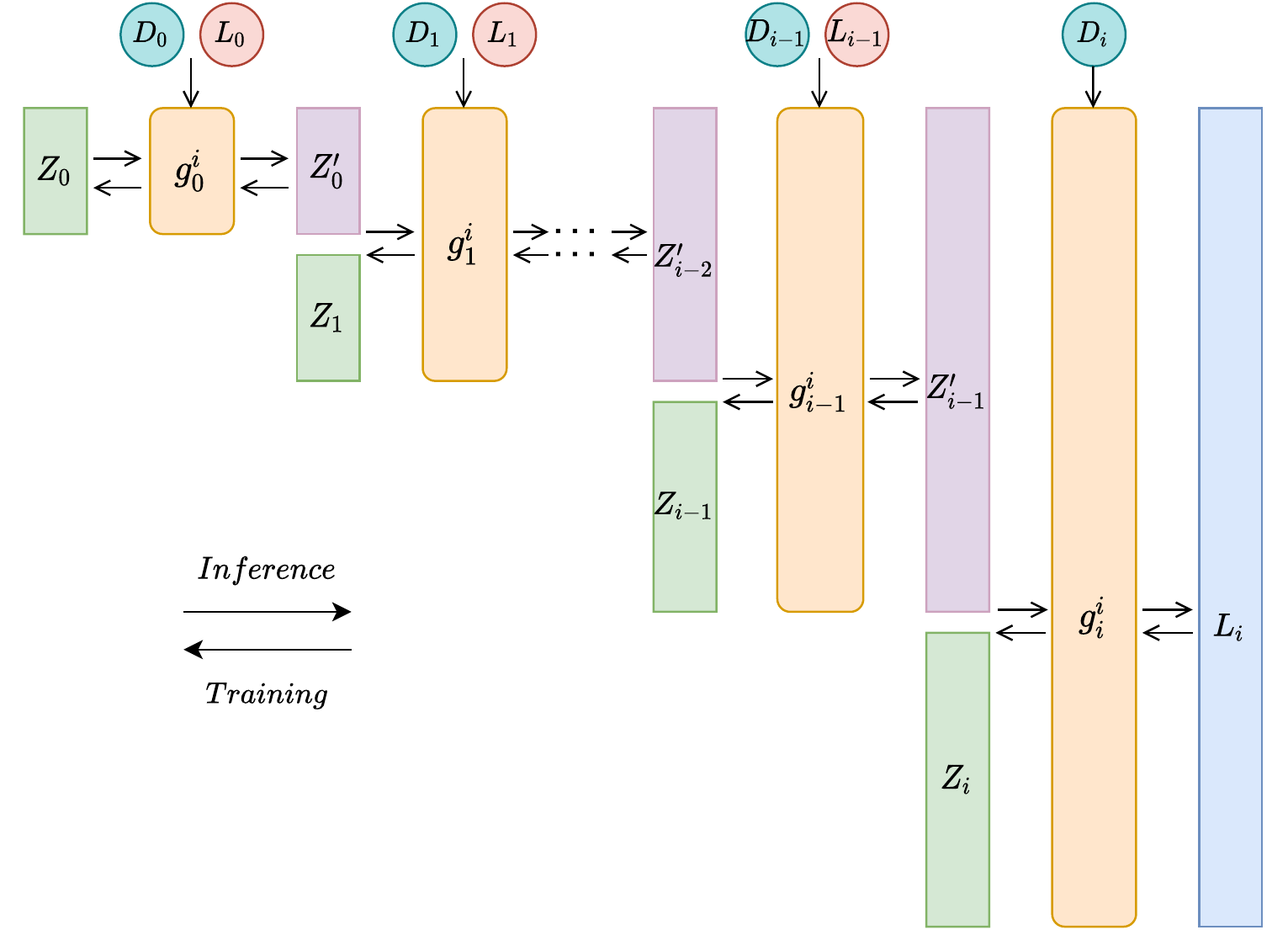}
\end{subfigure}
\caption{Left: unconditional normalizing flow architecture used to encode conditioning and conditioned images, denoted by $Y_n = Y$ and $W_n = W$ respectively, into a sequence of hierarchical latent spaces. Right: design of the conditional transformation $G_{i}^\theta$ that models the $i^{th}$ autoregressive component. The index of the flow $i$ is omitted in both the transformed latent spaces $Z_j$ and the intermediate latent spaces $Z_j^{\prime}$ for simplicity.}
   	\label{fig:high_level_design_conditional}
\end{figure}

We propose to estimate each autoregressive component $P(L_i|\tilde{L}_{i},\tilde D_{i+1})$ using a multi-scale conditional normalizing flow architecture. We define a sequence of latent spaces $\tilde Z_{i+1} := [Z_{i}^{i},...,Z_0^{i}]$ of decreasing dimension and a parametrized transformation
\begin{align*}
G^\theta_i : \tilde Z_{i+1} \times  \tilde D_{i+1} \times  \tilde{L}_{i} \rightarrow L_{i}.
\end{align*}
The transformations $G^\theta_i$ are constructed by assembling multi-scale transformations $(g_j^i)_{j=0}^i$ defined as follows:
\begin{align*}
&g_0^i : Z_0^i \times D_0 \times L_0 \rightarrow Z_{0}^{\prime i} \\   
 &  g_{j}^{i}:[Z_{j-1}^{\prime i}, Z_{j}^{i}] \times D_{j} \times  L_{j} \rightarrow Z_{j}^{\prime i}\, \quad \text{for} \quad j=1,...,i-1\\
&g_{i}^{i}: [Z_{i-1}^{\prime i}, Z_{i}^i] \times D_{i} \rightarrow     L_{i}
\end{align*}
where $Z_{i-1}^{\prime i}, \ldots, Z_0^{\prime i} $ are intermediate latent spaces of decreasing dimension, $g_j^i$ is invertible as a function from $[Z_{j-1}^{\prime i}, Z_{j}^i]$ to $Z_{j}^{\prime i}$ and $g_0^i$  is invertible as function from $Z_0^i$ to $Z_{0}^{\prime i}$.
The transformation $G^\theta_i$ is then obtained by composing the functions  $g_{j}^{i}$ in the following way. Given the conditioning variables $(\de_j)_{j=0}^i \in \tilde D_{i+1}$ and $(\el_j)_{j=0}^{i-1} \in \tilde L_{i}$, a latent variable $\z_0 \in Z_0^i$ is transformed to $\z_0' \in Z_0^{\prime i}$ by function $g^i_0(\cdot ; \de_0, \el_0)$. 
Then $\z_0'$ and $\z_1$ are concatenated and inserted to $g^i_1(\cdot;\de_1, \el_1)$ which outputs $\z_1' \in Z_1^{\prime i}$. This process continues as implied up until the $i^{th}$ level whose output is $\el_i = g_i^i([\z_{i-1}', \z_i]; \de_i) \in L_{i}$. A schematic description of $G^\theta_i$ is presented in the right diagram of Figure \ref{fig:high_level_design_conditional}. 

Denote by $f_{j}^{i}$ the inverse of $g_{j}^{i}$ for fixed conditioning variables in $D_j$ and $L_j$
\begin{align*}
&f_0^i : Z_0^{\prime i} \times D_0 \times L_0 \rightarrow Z_0^i \\   
 &  f_{j}^{i}:  Z_{j}^{\prime i} \times D_{j} \times  L_{j} \rightarrow [Z_{j-1}^{\prime i}, Z_{j}^i]\, \quad \text{for} \quad j=1,...,i-1\\
&f_{i}^{i}:    L_{i} \times D_{i} \rightarrow  [Z_{i-1}^{\prime i}, Z_{i}^i]
\end{align*}
and by $F_i^\theta$ the inverse of $G_i^\theta$ as a function from $\tilde Z_{i+1}$ to $L_i$ obtained by composing the functions $f^i_j$.
We model each $f_{j}^{i}$ by adopting the conditional flow design of \citet{SRFLOW}. More specifically, we initially use the same squeeze layer, which is followed by two transition steps and $K$ conditional flow steps. Each transition step consists of an actnorm layer followed by $1\times1$ invertible convolution layer. Each conditional flow step consists of an actnorm layer followed by $1\times1$ invertible convolution, which is followed by an affine injector and an affine coupling layer. We use from 8 to 16 conditional flow steps depending on the difficulty of the image translation task.

\subsection{Maximum log-likelihood estimation and training}
Here we use the constructed transformations $F_i^\theta$ to parametrize each autoregressive component $P(L_i|\tilde{L}_{i},\tilde D_{i+1})$ and, together with the unconditional normalizing flows $R^{\theta}$ and $T^{\theta}$, use them to estimate $P(W|Y)$.
With this aim, for every latent space $Z_j^i$ we choose as prior a multivariate normal distribution, whose density we denote by $\mathcal{N}(\z_j^i; \textbf{0}, \textbf{I})$.
Using the Bayes' rule, the change of variables formula for probability density functions, the factorization in \eqref{eq:factorization} and the chain rule for composition of functions, we parametrize the log-density of the conditional distribution $P(W|Y)$ by

\begin{align}
\log p^\theta_{W|Y}(\boldsymbol{w} | \y)   & = \sum_{i=1}^{n}\log{\Bigg|\text{det}\frac{\partial T_i^{\theta}(\boldsymbol{w}_{n-i+1})}{\partial \boldsymbol{w}_{n-i+1}}\Bigg|} \nonumber \\ 
& + \sum_{i=1}^{n-1} \sum^{i-1}_{j=0} \left[\log \mathcal{N}(\z_j^i;\textbf{0},\textbf{I})+\log \Bigg|\text{det}\left(\frac{\partial f^i_{j}(\z_j^{\prime i};\el_j,\de_{j})}{\partial \z_j^{\prime i}}\right)\Bigg|\right] \nonumber\\
& + \sum_{i=0}^{n-1}  \left[\log \mathcal{N}(\z_i^i;\textbf{0},\textbf{I})+\log \Bigg|\text{det}\left(\frac{\partial f^i_{i}(\el_{i};\de_{i})}{\partial \el_{i}}\right)\Bigg|\right]\label{eq:obj}
\end{align} 

where $\el_i \in L_i$, $\tilde{\el}_{i} \in \tilde L_{i}$ and $\widetilde{\de}_{i+1} \in \tilde D_{i+1}$ implicitly depend on $\boldsymbol{w}$ and $\y$ through the normalizing flows $R^\theta$ and $T^\theta$. We refer the reader to Section \ref{sec:derivations} in the Appendix for a detailed derivation \eqref{eq:obj}.

We train our CAFLOW model by minimizing the following training objective \begin{align*}
\log p^\theta_{W|Y}(\boldsymbol{w} | \y) + \lambda \log p^\theta_{Y}(\y),
\end{align*}
where $p^\theta_{Y}(\y)$ is estimated using the unconditional normalizing flow $R^\theta$ as in Subsection \ref{subsec:normalizing}. The parameter $\lambda$ acts as a regularizer during training. It can be interpreted as an interpolation parameter between the conditional distribution $P(W|Y)$ (when $\lambda = 0$) and the joint distribution $P(Y,W)$ (when $\lambda = 1$).

\subsection{Inference}

The standard way to perform inference using a trained CAFLOW model is the following: 
\begin{enumerate}
\item Calculate the conditional encodings $\de_{n-1},...,\de_0 \in \tilde D_n$ by passing the conditioning image through the multi-scale flow $R^\theta$.
\item Sample latent variables $\z_i^j$ from $\mathcal{N}(0,\tau^2)$, where $\tau$ denotes the sampling temperature.
\item Calculate the output image latent variables $\el_0,...,\el_{n-1} \in \tilde L_n$ by applying the transformations $G^\theta_i$ sequentially from $G^\theta_0$ to $G^\theta_{n-1}$.
\item Finally, convert the output image latents $\el_0,...,\el_{n-1}$ to the output image by passing them through the reverse flow $(T^\theta)^{-1}$.
\end{enumerate}

We have observed that using a sampling temperature $\tau$ less than $1$ typically leads to significant improvement in the quality of the output image at the expense of diversity. 

\begin{figure}[h!]
    \centering
    \includegraphics[width=\textwidth]{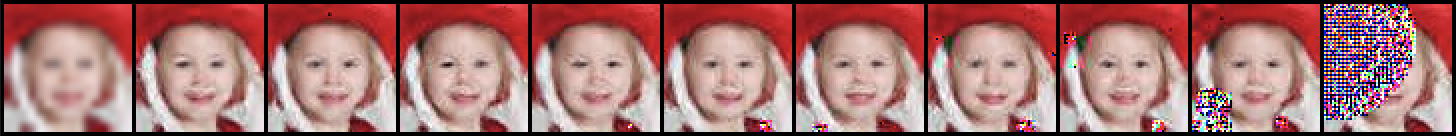}
    \caption{10 super-resolved versions of the LR image in decreasing conditional log-likelihood order.}
    \label{fig:decreasingloglikelihood}
\end{figure}

Our framework can act both as conditional generator and as a conditional likelihood estimator. Figure \ref{fig:decreasingloglikelihood} shows ten super-resolved versions of the low resolution image in decreasing order of conditional log-likelihood. We leverage the conditional likelihood estimation to automatically select the best generated samples and disregard bad samples. 

We have observed that CAFLOW typically assigns to the ground-truth image higher conditional log-likelihood than any generated sample. Such observation suggests that CAFLOW is a powerful conditional likelihood estimator, and that better samples can be obtained by the use of an optimisation algorithm, which searches for samples with high conditional likelihood. However, we used a simpler inference method in this work: keep the best $N$ out of $M$ generated samples based on the conditional log-likelihood. We empirically found that $M$ should be larger for higher sampling temperatures.

\section{Experiments}\label{sec:numerics}
In this section we evaluate CAFLOW on three image-to-image translation tasks: image super-resolution, image colorization and image inpainting. We refer the reader to section \ref{sec:details of experiments} in the Appendix for more details about the reported experiments and to section \ref{sec:extended visual results} for extended visual results.

\subsection{Image Super-resolution}
\begin{figure}
    \centering
    \includegraphics[width=.7\textwidth]{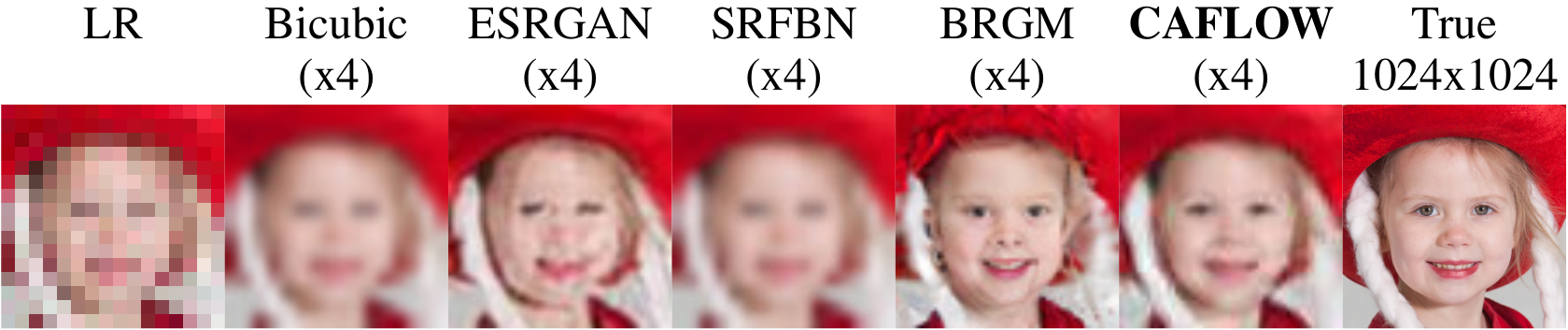}
    \caption{Qualitative evaluation on FFHQ 4x super-resolution of 16x16 resolution images. Left column shows the low resolution input, while the right column shows the true high-quality image. ESRGAN and SRFBN show clear distortion and blurriness. BRGM generates a clear image, but fails to recover the true image. Our method generates the closest image to the ground truth albeit slightly less clear than BRGM.}
    \label{SR visuals paper}
\end{figure}

We evaluate the ability of CAFLOW to perform image super-resolution on the FFHQ dataset \cite{karras2019style} (Creative Commons BY-NC-SA 4.0). We resized the original images to $16\times16$ and $64\times 64$ patches and trained the model for x4 super-resolution. The quantitative evaluation is performed using the LPIPS and RMSE scores on 100 unseen images. For inference, we used $\tau=0.5$ and kept the sample with the highest conditional log-likelihood out of ten generated samples. A comparison with state-of-the-art methods is shown in Table \ref{comparisonBRGM}. We present visual results in Figure \ref{SR visuals paper} and refer the reader to the supplementary material for more examples.

\newcommand{\lpi}{LPIPS}
\newcommand{\rms}{RMSE}

\begin{table}[h!]
\centering
\caption{Quantitative evaluation of (x4) super-resolution on FFHQ $16^2$. We report LPIPS/RMSE scores for each method. Lower scores are better.}\label{comparisonBRGM}
\label{eval-super-resolution}
\setlength{\tabcolsep}{3pt}

\begin{tabular}{l|ccccc}
 Dataset &    \textbf{CAFLOW} &   BRGM     &   ESRGAN    &   SRFBN  &  BICUBIC \\

\midrule
   FFHQ $16^2$ &  \textbf{0.08}/\textbf{17.56} & 0.24/25.66  &  0.35/29.32 &  0.33/22.07 & 0.34/20.10 \\

\end{tabular}
\end{table}

Our method outperforms all the other super-resolution methods based on both metrics. It is slightly inferior to BRGM in terms of perceptual quality but it is significantly better in terms of fidelity which is reflected in the quantitative evaluation.

\subsection{Image Colorization}
\begin{figure*}[h!]
\begin{center}
\setlength{\tabcolsep}{5pt}
\begin{tabular}{ccc}
CAFLOW & CINN & ColorGAN  \\
\includegraphics[height=3.2cm]{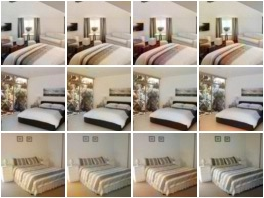}&
\includegraphics[height=3.2cm]{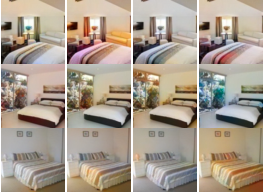}&
\includegraphics[height=3.2cm]{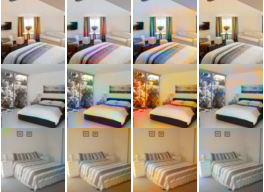}\\

\end{tabular}
\caption{Qualitative evaluation: Four colorizations proposed by CAFLOW, CINN and ColorGAN for three test images. ColorGAN generates unrealistically diverse colorizations with significant color artifacts (for example a yellow region on a white wall). CINN generates more realistic less diverse colorizations with less pronounced color artifacts compared to ColorGAN, which is reflected in the improved FID score. Finally, CAFLOW generates even more realistic and less diverse colorizations than CINN with even rarer color artifacts, which is more representative of the data distribution according to the FID score.}\label{colorisation paper visuals results}
\label{ffhq-super-resolution}
\end{center}
\end{figure*}

To our knowledge, diverse image colorization has been addressed by conditional flows \citep{ardizzone2019guided}, conditional GANs \citep{colorGAN} and recently score-based models \citep{scorebased}. 
We trained the model on $10\%$ of the LSUN bedroom $64\times 64$ training dataset \citep{yu2015lsun},  a popular dataset for image colorization. For inference, we used $\tau=0.85$ and kept the best out of ten generated samples for each test image using the conditional log-likelihood. We report the performance of the model in Table \ref{colorisation comparison}. We use the FID score to compare our method against the cINN and ColorGAN, which have been trained on the full dataset. We show visual results for all methods in Figure \ref{colorisation paper visuals results}. 
We did not include \citep{scorebased} in the comparison because it was trained on higher resolution images.

\begin{table}[h!]
\centering

\caption{Quantitative evaluation of colorization on LSUN BEDROOM $64\times 64$ dataset. We report FID score for each method. Lower scores are better. }
\label{colorisation comparison}
\setlength{\tabcolsep}{3pt}

\begin{tabular}{l|ccccc}
 Metric &    \textbf{CAFLOW} &   CINN     &   ColorGAN   \\

\midrule
  FID &  \textbf{18.15} & 26.48 &  28.31 \\

\end{tabular}
\end{table}

Our model outperforms both methods on image colorization based on the FID metric. We calculated the FID score by generating one conditional sample for each test image, so that our evaluation is identical with the evaluation of the other methods. Moreover, we calculated the FID score by generating 5 samples for each test image, which yielded an improved FID score ($\textbf{16.73}$). We suggest that this protocol be adopted by methods which address diverse image colorization in the future. 

\subsection{Image Inpainting}

\begin{figure}[h!]
    \centering
    \includegraphics[width=.7\textwidth]{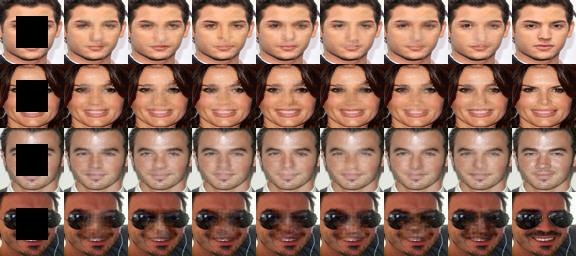}
    \caption{Different inpaintings proposed by CAFLOW with $\tau=0.5$. Ground truth on the right.}
    \label{qualitative performance inpainting}
\end{figure}

We evaluated the performance of CAFLOW on image inpainting by removing central masks covering $25\%$ of the centrally cropped human face of the CelebA dataset \citep{liu2015deep}. We compare the performance of the model with the conditional flow \citep{cGLOW} on the same task using the PSNR metric, see Table \ref{quantitative evaluation inpainting}. We show inpainting examples in Figure \ref{qualitative performance inpainting}.

Our model outperforms \citep{cGLOW} based on the PSNR metric and on qualititative performance. CAFLOW generates realistic images by blending smoothly the conditioning with the synthesized part of the image in contrast to \citep{cGLOW} which generates overly smooth synthesized parts which do not blend well with the surrounding image. Both methods fail in faces which wear sunglasses or face sidewise. We believe that this is attributed to the small number of such training examples.

\begin{table}[h!]
\centering
\caption{Quantitative evaluation of inpainting on the CelebA dataset. We report PSNR and LPIPS scores for each method.}\label{quantitative evaluation inpainting}
\setlength{\tabcolsep}{3pt}
\begin{tabular}{l|cccc}
 Method  &  PSNR$\uparrow$  &   LPIPS$\downarrow$    \\

\midrule
  \textbf{CAFLOW} &  \textbf{26.08} & 0.06  \\
  \citet{cGLOW} & 24.88 & -
    
\end{tabular}
\end{table}

\section{Limitations}\label{sec:limitations}

The main modeling limitation of the framework is attributed to limited expressivity of normalizing flows. Each transformation of a normalizing flow has to be invertible with a tractable calculation of the determinant of the jacobian. This limits the representational power of normalizing flows. However, Wu et al. \citep{wu2020stochastic} claim to overcome expressivity limitations of normalizing flows by combining deterministic invertible functions with stochastic sampling blocks. Therefore, incorporating their proposed stochastic blocks in our conditional autoregressive flows can potentially lead to significant performance improvement. 

Another limitation of the framework is that it can become computationally expensive for high resolution images where typically more scales are used for the hierarchical latent decomposition of the images. A possible solution to this problem is the use of weight sharing among the conditional autoregressive components as described in section \ref{sec:weight sharing strategy} in the Appendix.

\section{Conclusion}

We have introduced conditional auto-regressive flows, coined CAFLOW, which combine auto-regressive modeling with conditional normalizing flows for image-to-image translation. CAFLOW is an efficient conditional image generator and conditional likelihood estimator able to cross-correlate information at different scales, improving on the expressivity of former conditional flow architectures. We demonstrate its efficiency as conditional generator on standard image-to-image translation tasks such as image super-resolution, image colorization and image inpainting. We find that, in the above-mentioned tasks, CAFLOW achieves better performance than standard conditional flows and conditional GANs. Moreover, we demonstrate its efficiency as conditional likelihood estimator by showing that the model can disregard bad samples and keep the best samples based on the conditional log-likelihood.

\begin{ack}
We thank Lynton Ardizzone for informing us about efficient ways to train normalizing flows. We also thank Razvan Marinescu and You Lu for providing help with practical details of the evaluation procedure.

GB acknowledges support from GSK. 
MC acknowledges support from the Royal Society (Newton International Fellowship NIF\textbackslash R1\textbackslash 192048 Minimal partitions as a robustness boost for neural network classifiers).
CE acknowledges support from the Wellcome Innovator Award RG98755. ZK acknowledges funding from the Royal Society (Industry Fellowship), Wellcome Trust (205067/Z/16/Z), Alan Turing Institute and the Alzheimer’s Drug Discovery Foundation.
CBS acknowledges support from the Philip Leverhulme Prize, the Royal Society Wolfson Fellowship, the EPSRC grants EP/S026045/1 and EP/T003553/1, EP/N014588/1, EP/T017961/1, the Wellcome Innovator Award RG98755, the Leverhulme Trust project Unveiling the invisible, the European Union Horizon 2020 research and innovation programme under the Marie Skodowska-Curie grant agreement No. 777826 NoMADS, the Cantab Capital Institute for the Mathematics of Information and the Alan Turing Institute.

\end{ack}

\bibliographystyle{plainnat}
\bibliography{bibliography}

\clearpage

\appendix
\label{appendix}

\section{Architecture}

In this section, we provide more details about the architecture of the unconditional multi-scale flows and the conditional auto-regressive multi-scale flows which comprise our modelling framework shown in Figure \ref{fig:high_level_design_conditional}.

\subsection{Unconditional multi-scale flows}
The sequence of transformations in each scale are:

\begin{enumerate}
    \item \textbf{Dequantization layer} (only in the first scale). Each pixel of each image channel admits an integer value from $0$ to $255$. If we do not modify those values, we model a high dimensional distribution in discrete space. However, normalizing flows rely on the rule of change of variables which is naturally defined in continuous space. Therefore, training a normalizing flow on the raw images will result in a model which which will place arbitrarily high likelihood on a few RGB values. This behavior leads to catastrophic destabilization of the training procedure and very poor performance. To remedy this problem, we convert the discrete space to continuous space by dequantizing each pixel value by resampling it from the distribution $q(u_{deq}|u)=\mathcal{U}([u,u+1))$, where $\mathcal{U}([u,u+1))$ is the uniform distribution over the interval $[u,u+1)$. This is called \textbf{uniform dequantization} and it is used by the majority of normalizing flow models in the literature. Dequantization effectively represents images as hypercubes. Modelling such sharp borders is challenging for a normalizing flow as it relies on smooth transformations. For this reason, \citet{ho2019flow++} proposed to learn the dequantization distribution $q(x_{deq}|x)$, where $x$ is the quantized and $x_{deq}$ the dequantized image using a conditional normalizing flow. This is called \textbf{variational dequantization}. We noticed that using variational dequantization results in a significant increase in performance. For this reason, we decided not to use variational dequantization in the experiments where we compare against other conditional flows, because this would defeat the purpose of fair comparison. We used variational dequantization in the experiments where we compare against methods which do not rely on normalizing flows.
    
    \item \textbf{Squeeze layer}. We squeeze the tensor in the channel dimension effectively halving the spatial resolution: the shape of the tensor is transformed from $(B, C, H, W)$ to $(B, 4C, H/2, W/2)$, where $B$ is the batch size, $C$ the number of channels, $H$ the height and $W$ the width of the image.
    \item \textbf{Two transition steps}. Each transition step consists of an invertible normalization layer (ActNorm) followed by an invertible $1\times 1$ convolution layer. According to \citep{SRFLOW}, the transition step allows the network to learn a linear invertible interpolation between neighboring pixels after the application of the squeeze layer. If the transition steps are not used, the squeeze layer may lead to checkerboard artifacts in the reconstructed image because it is exclusively based on pixel-reordering.
    \item \textbf{K flow steps}. Each flow step consists of an invertible normalization layer, invertible $1\times 1$ convolution layer and an affine coupling layer in that order. The scale and the bias in the affine coupling layer are computed by a simple convolutional neural network which performs three sequential convolutions with a $3\times 3$, $1\times 1$ and $3\times 3$ convolutional kernels respectively. The number of kernels in each convolutional layer is chosen to be $64$.
    \item \textbf{Split layer}. We split off half of the channels before the squeeze layer of the next scale. Therefore, the next scale transforms only the half part of the tensor. This motivates the modelling of variations in different resolutions and ultimately the hierarchical latent space decomposition of images using multi-scale normalizing flows. Empirically, it has been observed that the first split variables typically encode noise patterns or image details, while the final split variables encode higher level information.
    
\end{enumerate}

\clearpage

\subsection{Conditional multi-scale flows}

Each scale of a conditional flow $F_i^{\theta}$ contains the following transformations in order:

\begin{enumerate}
    \item \textbf{Squeeze layer}.
    \item \textbf{Two transition steps}.
    \item \textbf{M Conditional flow steps}. We adopt the conditional flow step design used by \citep{SRFLOW}: \begin{enumerate}
        \item Invertible normalization (ActNorm).
        \item Invertible $1\times 1$ convolution.
        \item Affine injector.
        \item Conditional affine coupling layer.
    \end{enumerate}
    \item \textbf{Split layer}.
\end{enumerate}

We calculate the scale and the bias of the affine transformations in the affine injector and the conditional affine coupling layer using the same convolutional neural network as the one we used for the unconditional flows. The only difference is that we use $32$ instead of $64$ kernels.

\section{Details of experiments}\label{sec:details of experiments}
We used the same learning rate scheduling system for all experiments. Initially, we increase the learning rate linearly from $0$ to the target learning rate (usually $10^{-3}$) in the first $500$ iterations. Then, we use the pytorch STEPLR learning rate scheduler with value $\gamma = 0.999$. When the training curve levels-off (around $70$K iterations), we reduce the learning rate by a factor of $10$ until final convergence. This reduction provides a final performance boost. We do not repeat that reduction because it only results in overfitting. 

Moreover, we used exponential moving average (EMA) with a rate equal to $0.999$ in all experiments. We empirically found that EMA consistently provides a small increase in performance. Finally, we clipped the norm of the computed gradients to $1$, because it improved training stability with no noticeable compromise on performance.

\subsection{Image super-resolution}
We used $3$ scales, $16$ flow steps in each level of the conditioning flow $R^\theta$, $32$ flow steps in the each level of the conditioned flow $T^\theta$ and $12$ conditional flow steps in each level of each conditional flow $F_i^{\theta}$. We set the regularization constant $\lambda = 0.01$. Moreover, we used variational dequantization implemented with a conditional flow of $4$ conditional flow steps, because we compared our method against methods which are not based on flows. We trained our model for $4.5$ days on a single NVIDIA TESLA P100 GPU on $68$K images of the FFHQ dataset with a target learning rate of $10^{-3}$. We used $1000$ images for validation and $100$ images for testing.

\subsection{Image colorization}

We used $3$ scales, $24$ flow steps in each level of the conditioning flow $R^\theta$, $24$ flow steps in the each level of the conditioned flow $T^\theta$ and $12$ conditional flow steps in each level of each conditional flow $F_i^{\theta}$. We set the regularization constant $\lambda = 0.01$. We used uniform dequantization as we intended to compare our method with \citep{ardizzone2019guided}, which is a conditional flow. We trained the model on $300$K images of the lsun bedroom dataset (this accounts for $10\%$ of the full dataset) on a single NVIDIA TESLA P100 GPU for 3 days with a target learning rate of $10^{-3}$. We used $1000$ validation images and $5000$ test images.

\subsection{Image inpainting}
We used $3$ scales, $30$ flow steps in each level of the conditioning flow $R^\theta$, $30$ flow steps in the each level of the conditioned flow $T^\theta$ and $16$ conditional flow steps in each level of each conditional flow $F_i^{\theta}$. We set the regularization constant $\lambda = 0.05$ and the target learning rate to $10^{-4}$, because we faced stability issues with smaller values of $\lambda$ and greater values of the target learning rate. We trained the model on $195$K images of the CelebA dataset for $5$ days on a single NVIDIA TESLA P100 GPU. We used 1000 images for validation and $2000$ images for testing. We used the same preprocessing as  \citet{cGLOW} and uniform dequantization.

\section{Derivations}\label{sec:derivations}

In this section we provide a detailed derivation of \eqref{eq:obj}, which is part of the training objective.
\begin{equation*}
\begin{aligned}
     \log p^\theta_{W|Y}(\boldsymbol{w} | \y)=& \log \Big|\text{det}\Bigg(\frac{\partial T^\theta(\boldsymbol{w})}{\partial \boldsymbol{w}}\Bigg)\Big| + \log{p^\theta_{\tilde{L}_n|\tilde{D}_n}(\tilde{\el}_n|\widetilde{\de}_n)} \\ 
    =& \sum_{i=1}^{n}\log{\Bigg|\text{det}\frac{\partial T_i^{\theta}(\boldsymbol{w}_{n-i+1})}{\partial \boldsymbol{w}_{n-i+1}}\Bigg|}+ \log{\prod_{i=0}^{n-1}p^\theta_{L_i|\tilde{L}_{i},\tilde D_{i+1}}(\el_i|\tilde{\el}_{i},\widetilde \de_{i+1})} \\
    =& \sum_{i=1}^{n}\log{\Bigg|\text{det}\frac{\partial T_i^{\theta}(\boldsymbol{w}_{n-i+1})}{\partial \boldsymbol{w}_{n-i+1}}\Bigg|}+ \sum_{i=0}^{n-1}\log{p_{L_i|\tilde{L}_{i},\tilde D_{i+1}}^\theta(\el_i|\tilde{\el}_{i},\widetilde \de_{i+1})}.
\end{aligned}
\end{equation*}

The justification for the first line can be found in Section 3 of \citep{dual-glow}. We go from the first line to the second using the chain rule for the first term and by factorizing the conditional distribution under the dependency assumptions of our model for the second term. The calculation of the first term is straightforward, because we have chosen invertible components with a tractable calculation of the determinant of the Jacobian. We name the second term $B$ for further analysis.

\begin{equation*}
\begin{aligned}
    B =& \sum_{i=0}^{n-1}\log{p_{L_i|\tilde{L}_{i},\tilde D_{i+1}}^\theta(\el_i|\tilde{\el}_{i},\widetilde \de_{i+1})} \\
     =& \sum_{i=0}^{n-1}\log{p_{\tilde{Z}_{i+1}}(F_i^{\theta}(\el_i;\tilde{\el}_{i},\widetilde \de_{i+1}))}+\log{\Bigg|\text{det}\frac{\partial F_i^{\theta}(\el_i;\tilde{\el}_{i},\widetilde \de_{i+1})}{\partial \el_i}\Bigg|} \\
    =& \sum_{i=1}^{n-1} \sum^{i-1}_{j=0} \left[\log \mathcal{N}(\z_j^i;\textbf{0},\textbf{I})+\log \Bigg|\text{det}\left(\frac{\partial f^i_{j}(\z_j^{\prime i};\el_j,\de_{j})}{\partial \z_j^{\prime i}}\right)\Bigg|\right] \nonumber\\
& + \sum_{i=0}^{n-1}  \left[\log \mathcal{N}(\z_i^i;\textbf{0},\textbf{I})+\log \Bigg|\text{det}\left(\frac{\partial f^i_{i}(\el_{i};\de_{i})}{\partial \el_{i}}\right)\Bigg|\right]
\end{aligned}
\end{equation*}

The second line is simply obtained by using the change of variables formula. We go from the second line to the third by using the chain rule for the composition of functions (remember the definition of $F_{i}^{\theta}$ in \ref{Autoregressive conditional flow components}) and the assumption that the latent variables $Z_i^i,Z_{i-1}^i, ..., Z_0^i$ which comprise $\tilde{Z}_{i+1}$ are i.i.d. with distribution $\mathcal{N}(\textbf{0},\textbf{I})$. The sum is broken into two sums, because the $f_i^i$ functions are conditioned only on $\de_i$. Therefore, 

\begin{align}
\log p^\theta_{W|Y}(\boldsymbol{w} | \y)   & = \sum_{i=1}^{n}\log{\Bigg|\text{det}\frac{\partial T_i^{\theta}(\boldsymbol{w}_{n-i+1})}{\partial \boldsymbol{w}_{n-i+1}}\Bigg|} \nonumber \\ 
& + \sum_{i=1}^{n-1} \sum^{i-1}_{j=0} \left[\log \mathcal{N}(\z_j^i;\textbf{0},\textbf{I})+\log \Bigg|\text{det}\left(\frac{\partial f^i_{j}(\z_j^{\prime i};\el_j,\de_{j})}{\partial \z_j^{\prime i}}\right)\Bigg|\right] \nonumber\\
& + \sum_{i=0}^{n-1}  \left[\log \mathcal{N}(\z_i^i;\textbf{0},\textbf{I})+\log \Bigg|\text{det}\left(\frac{\partial f^i_{i}(\el_{i};\de_{i})}{\partial \el_{i}}\right)\Bigg|\right] \nonumber
\end{align} 

\clearpage

\section{Weight sharing}\label{sec:weight sharing strategy}

\begin{figure}[h!]
\centering
\begin{subfigure}{0.45\textwidth}
	\includegraphics[width=\textwidth]{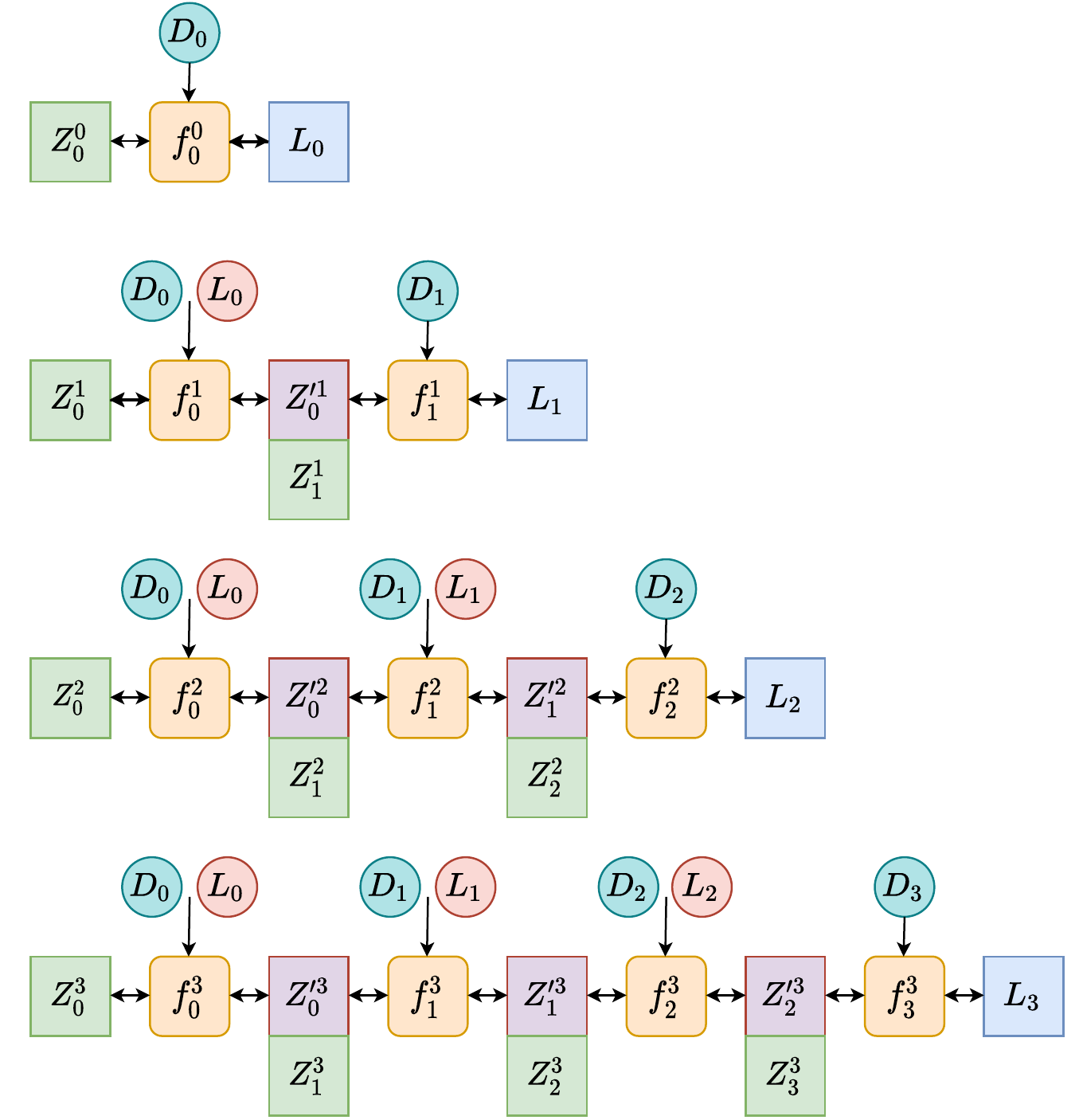}
\end{subfigure}%
\qquad 
\begin{subfigure}{0.47\textwidth}
		\includegraphics[width=\textwidth]{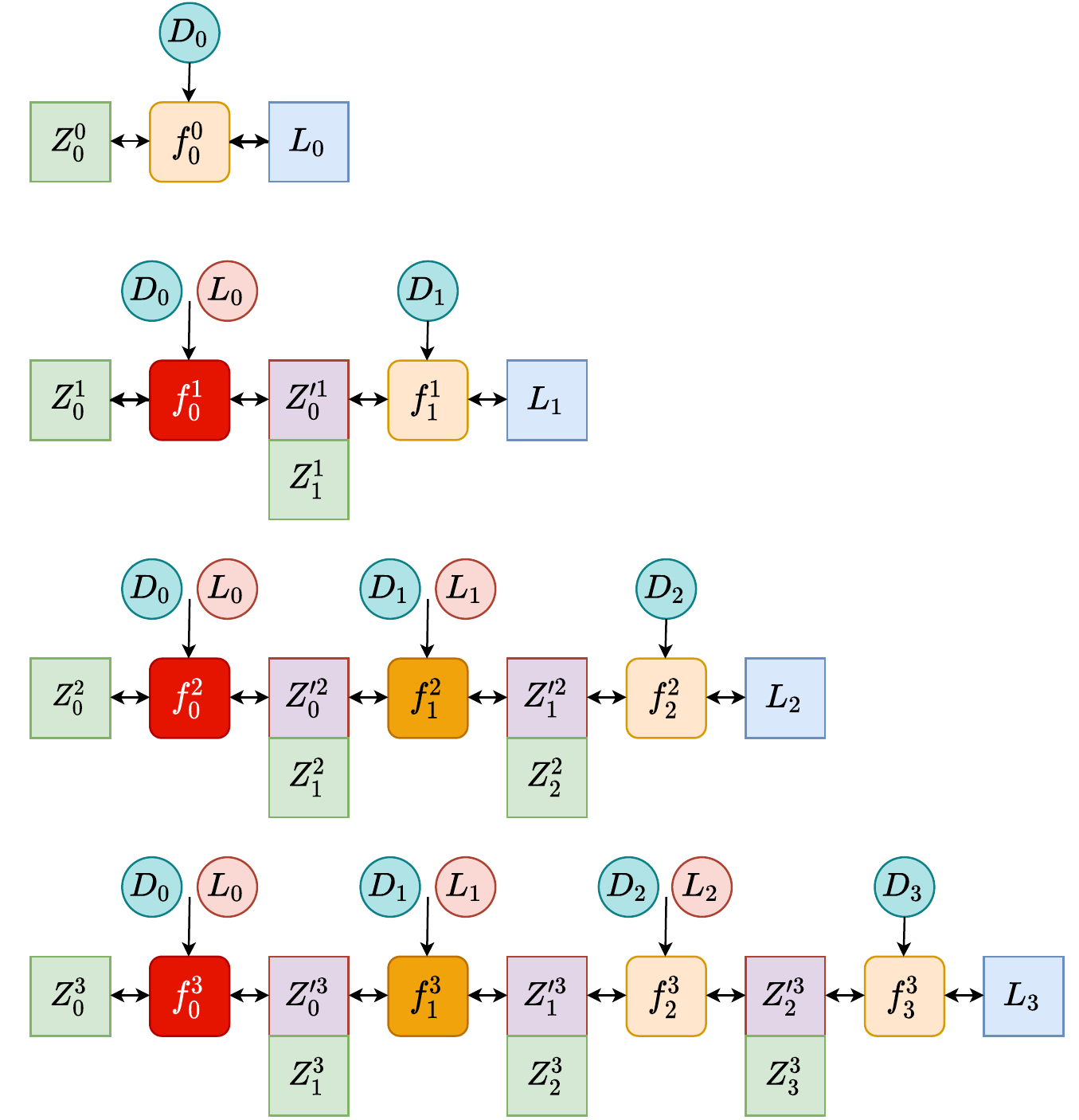}
\end{subfigure}
\caption{Conditional auto-regressive framework with four scales. Left: No weight sharing. Total number of functions: $n(n+1)/2$ where $n$ the number of scales. Right: Weight sharing (functions with the same vivid color share weights). Total number of functions: $2n-1$.}
   	\label{fig:framework-shared-unshared}
\end{figure}

We mentioned in the second paragraph of Section \ref{sec:limitations} that the framework can become computationally expensive for modelling conditional distributions of high resolution images, because more scales are needed for the latent decomposition of those images. We claimed that a possible solution to this is the sharing of certain weights between conditional auto-regressive flows of different levels. In particular, we propose that the weights of the networks $f_k^{k+1}, f_k^{k+2},...f_k^{n-1}$ are shared for all $k$ satisfying $0\leq k \leq n-2$. To make this clearer, we present two versions of the framework with four scales in Figure \ref{fig:framework-shared-unshared}: one with no sharing and one where we highlight the shared functions with the same color. Our choice to use the same weights for those functions is based on the fact that those functions share the same conditions and also operate on activations of the same shape at the same level down the flows.

We experimented with this modification for low resolution images ($64\times 64$ images), because we intended to compare it with the theoretical approach. We found that it can yield good performance albeit slightly inferior to the performance of the theoretical framework. Given that this modification reduces the number of internal $f_i^j$ functions from quadratic to linear and still performs well, we believe that it could be used successfully for modelling conditional distributions of high resolution images where more scales/levels $n$ are typically needed. We intend to explore this direction further in the future.

\clearpage
\section{Visual results}\label{sec:extended visual results}

\subsection{Image super-resolution}
\begin{figure}[h!]
    \centering
    \includegraphics[width=\textwidth]{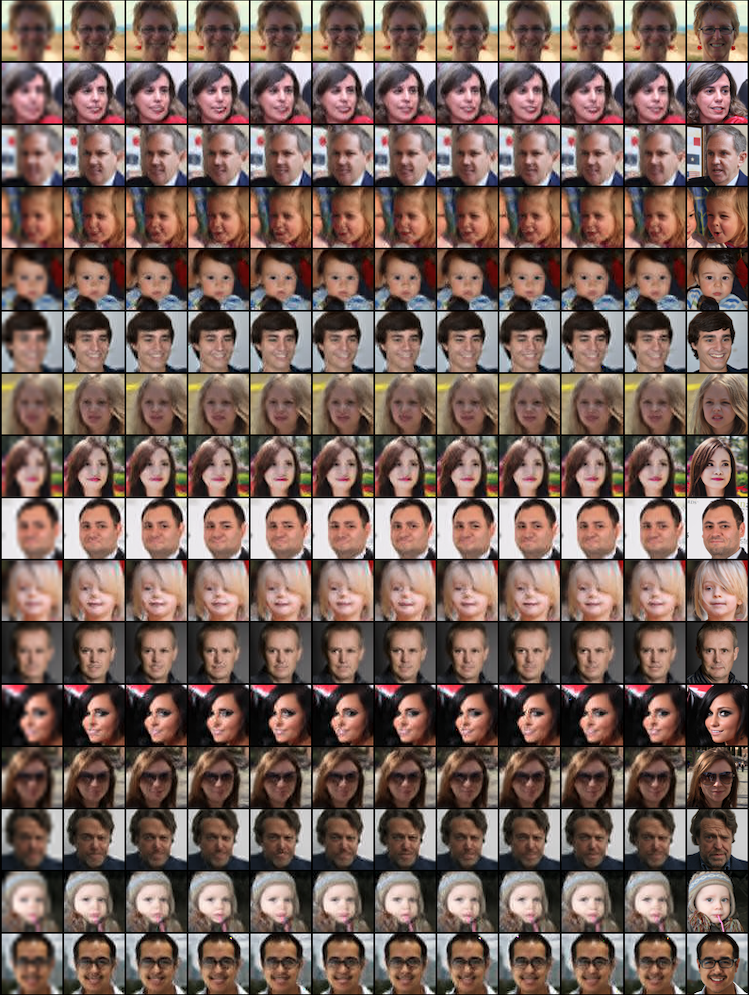}
    \caption{Image super-resolution on the FFHQ dataset. Left: LR bicubicly upsampled. Right: HR image. Middle: 10 super-resolved versions in decreasing conditional log-likelihood order from left to right. We sampled 20 super-resolved images for each LR image and we present the 10 images with the highest conditional log-likelihood. We used sampling temperature $\tau=0.5$.}
\end{figure}

\begin{figure}[h!]
    \centering
    \includegraphics[width=\textwidth]{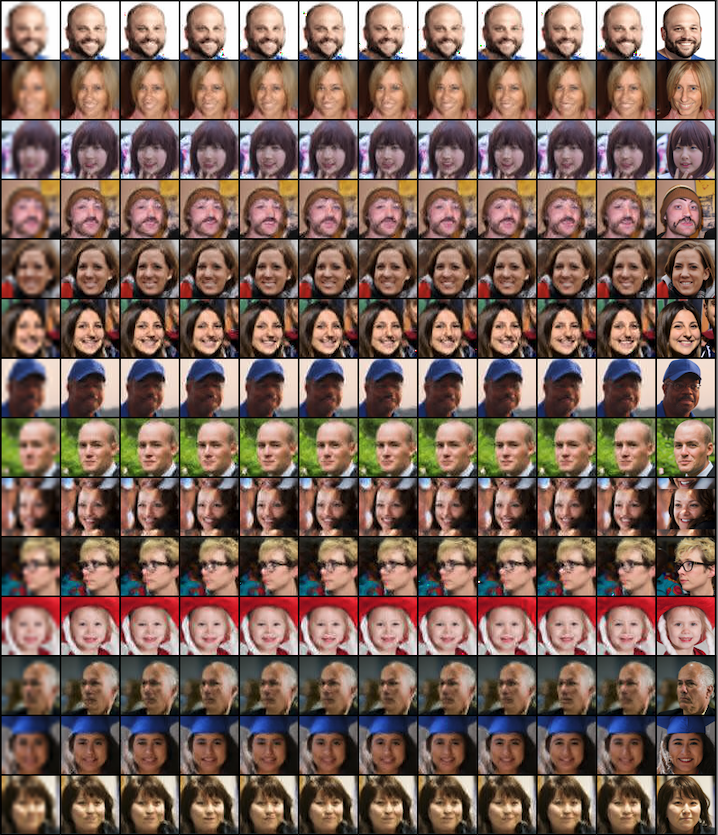}
    \caption{Image super-resolution on the FFHQ dataset. Left: LR bicubicly upsampled. Right: HR image. Middle: 10 super-resolved versions in decreasing conditional log-likelihood order from left to right. We sampled 20 super-resolved images for each LR image and we present the 10 images with the highest conditional log-likelihood. We used sampling temperature $\tau=0.55$.}
\end{figure}

\clearpage
\subsection{Image inpainting}
\begin{figure}[h!]
    \centering
    \includegraphics[width=\textwidth]{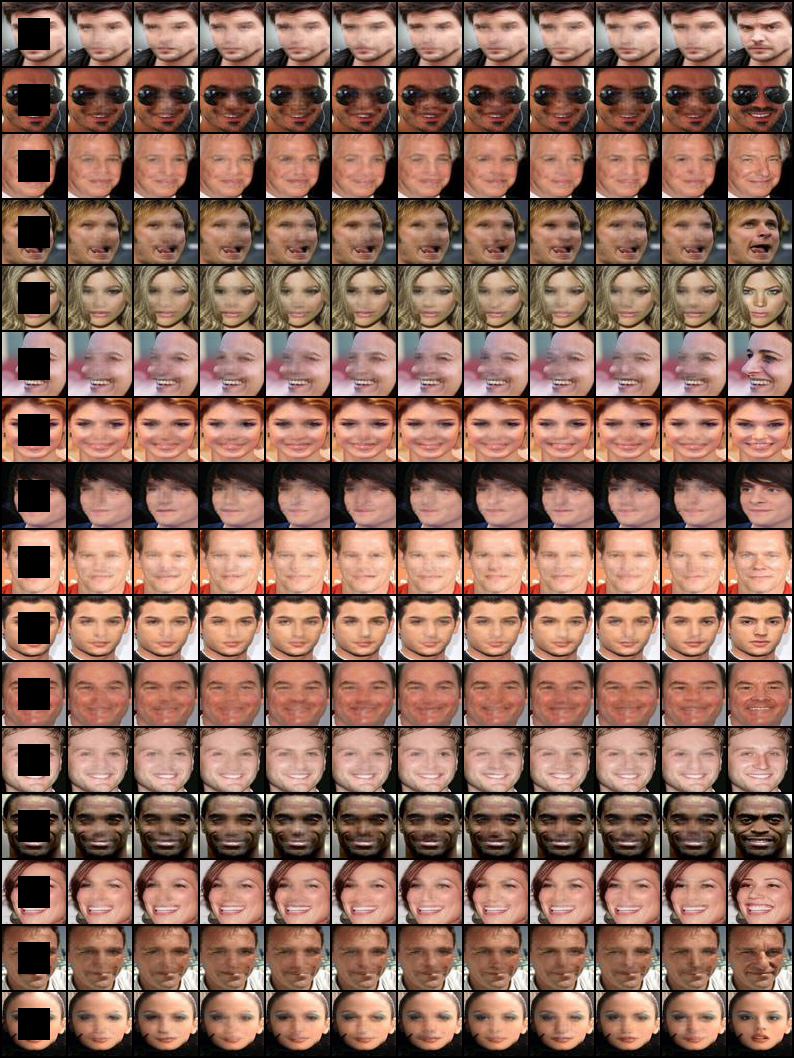}
    \caption{Image inpainting on the CelebA dataset. Left: Masked image. Right: Ground truth. Middle: 10 inpainted versions in decreasing conditional log-likelihood order from left to right. We sampled 30 inpainted images for each masked image and we present the 10 images with the highest conditional log-likelihood. We used sampling temperature $\tau=0.5$.}
\end{figure}

\begin{figure}[h!]
    \centering
    \includegraphics[width=\textwidth]{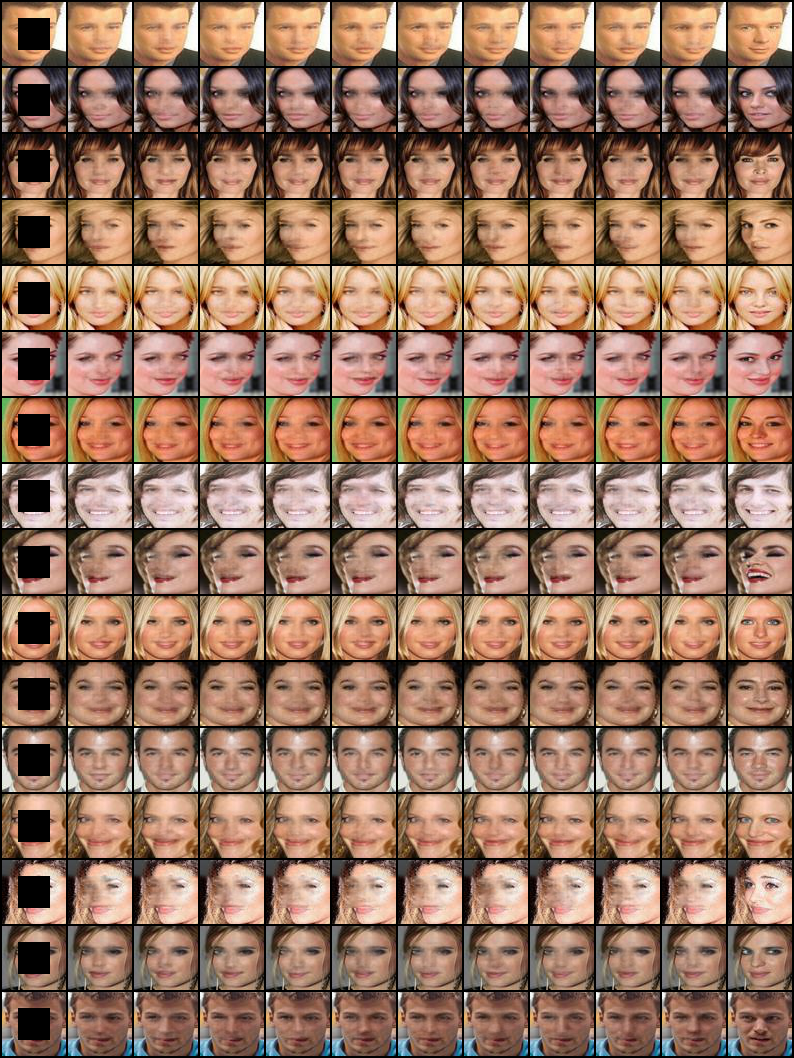}
    \caption{Image inpainting on the CelebA dataset. Left: Masked image. Right: Ground truth. Middle: 10 inpainted versions in decreasing conditional log-likelihood order from left to right. We sampled 30 inpainted images for each masked image and we present the 10 images with the highest conditional log-likelihood. We used sampling temperature $\tau=0.5$.}
\end{figure}
\clearpage
\subsection{Image colorization}
\begin{figure}[h!]
    \centering
    \includegraphics[width=\textwidth]{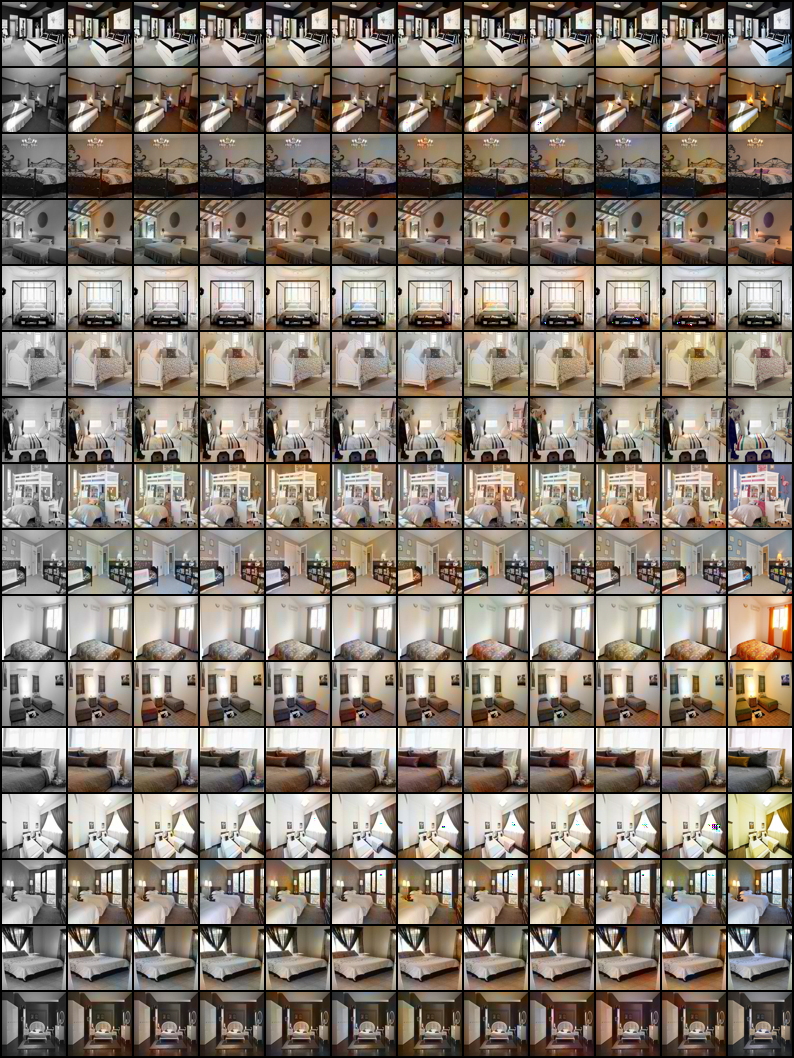}
    \caption{Image colorization on the LSUN BEDROOM dataset. Left: Grayscale image. Right: Ground truth. Middle: 10 colorized versions in decreasing conditional log-likelihood order from left to right. We sampled 25 colorized images for each greyscale image and we present the 10 images with the highest conditional log-likelihood. We used sampling temperature $\tau=0.85$.}
\end{figure}

\begin{figure}[h!]
    \centering
    \includegraphics[width=\textwidth]{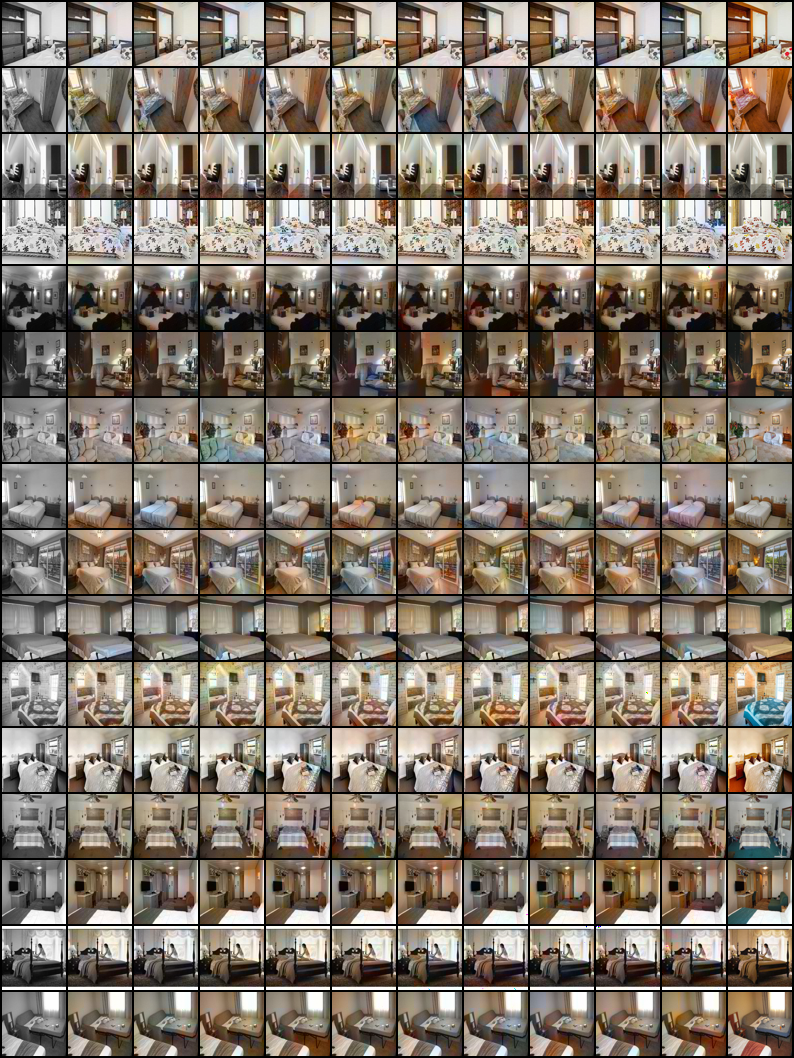}
    \caption{Image colorization on the LSUN BEDROOM dataset. Left: Grayscale image. Right: Ground truth. Middle: 10 colorized versions in decreasing conditional log-likelihood order from left to right. We sampled 25 colorized images for each greyscale image and we present the 10 images with the highest conditional log-likelihood. We used sampling temperature $\tau=0.85$.}
\end{figure}

\begin{figure}[h!]
    \centering
    \includegraphics[width=\textwidth]{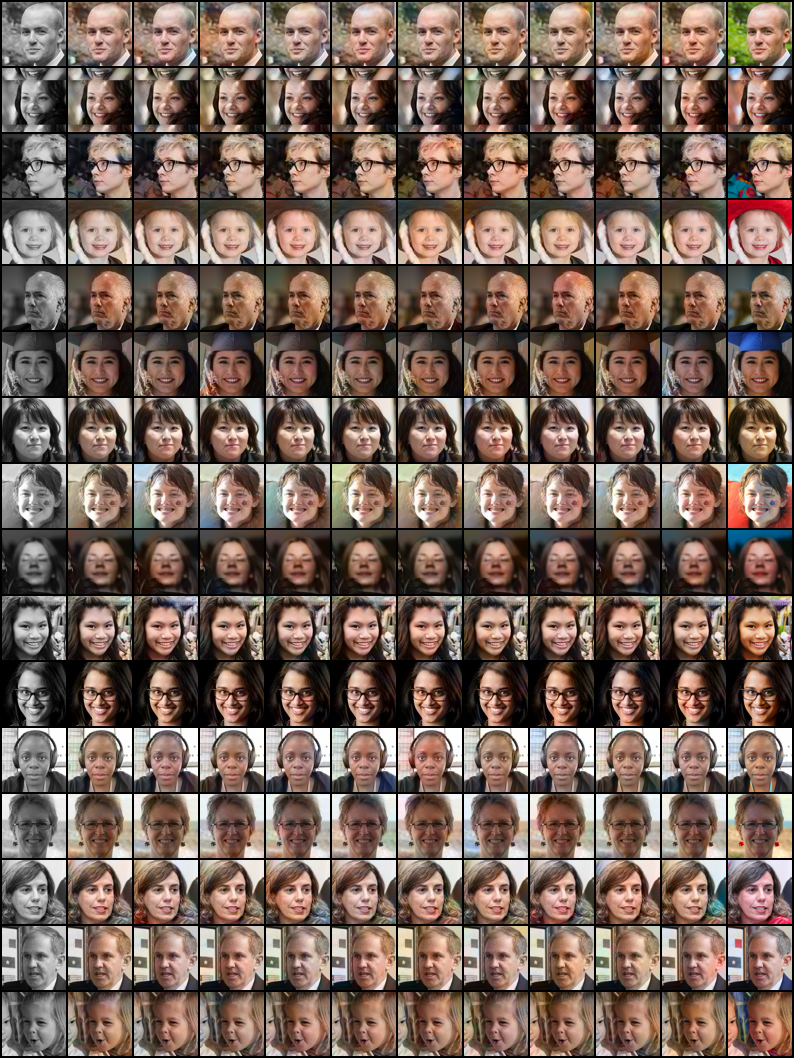}
    \caption{Image colorization on the FFHQ dataset. Left: Grayscale image. Right: Ground truth. Middle: 10 colorized versions in decreasing conditional log-likelihood order from left to right. We sampled 25 colorized images for each greyscale image and we present the 10 images with the highest conditional log-likelihood. We used sampling temperature $\tau=0.7$.}
\end{figure}

\begin{figure}[h!]
    \centering
    \includegraphics[width=\textwidth]{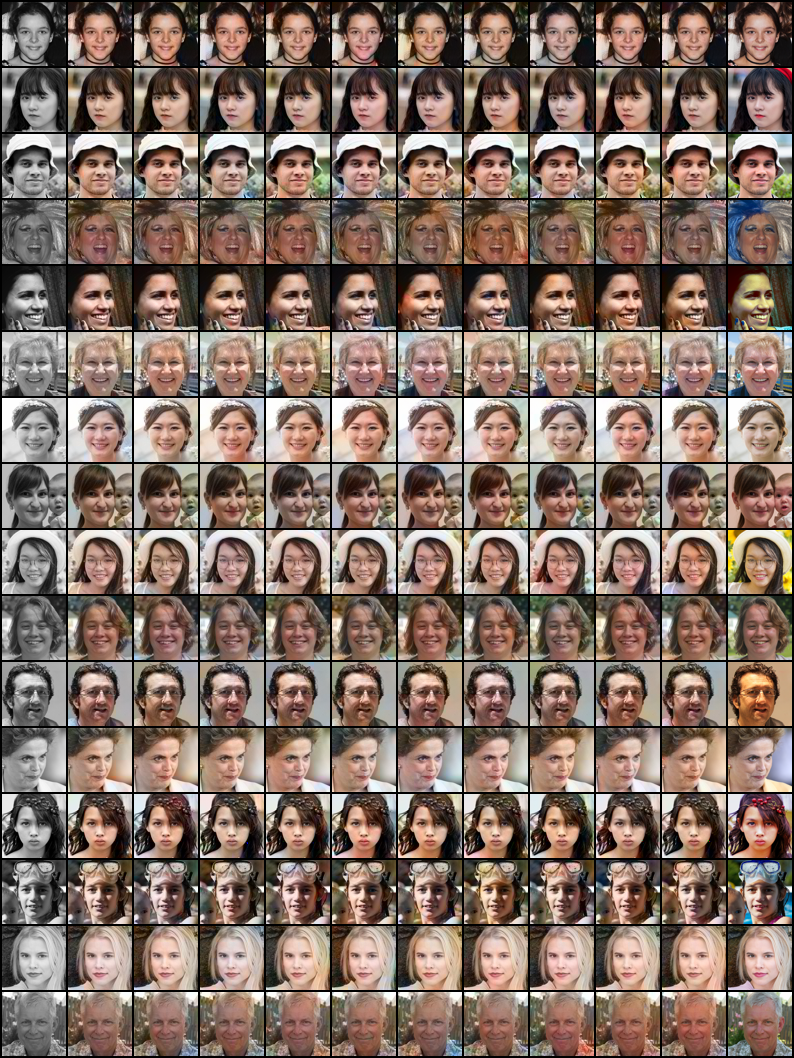}
    \caption{Image colorization on the FFHQ dataset. Left: Grayscale image. Right: Ground truth. Middle: 10 colorized versions in decreasing conditional log-likelihood order from left to right. We sampled 25 colorized images for each greyscale image and we present the 10 images with the highest conditional log-likelihood. We used sampling temperature $\tau=0.7$.}
\end{figure}

\clearpage

\subsection{Sketch to image synthesis}

\begin{figure}[h!]
    \centering
    \includegraphics[scale=0.5]{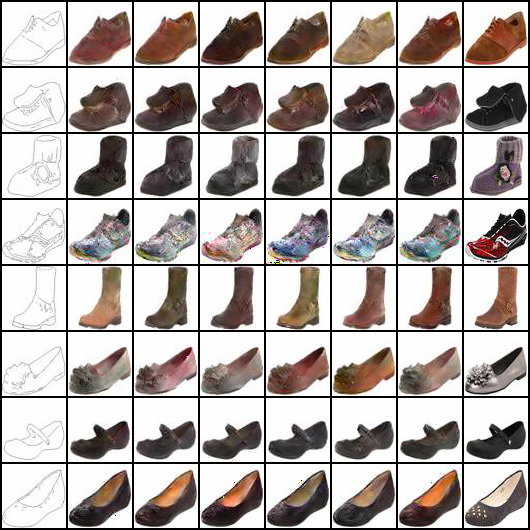}
    \caption{Sketch to image synthesis on the edges2shoes dataset \cite{isola2017image}. Left: Sketch. Right: Ground truth. Middle: 6 samples taken with sampling temperature $\tau=0.8$.}
\end{figure}

\end{document}